%% file: anonymous-submission-latex-2026.tex
\documentclass[letterpaper]{article} 
\usepackage{aaai2026}  
\usepackage{times}  
\usepackage{helvet}  
\usepackage{courier}  
\usepackage[hyphens]{url}  
\usepackage{graphicx} 
\usepackage{natbib}  
\usepackage{caption} 
\usepackage{algorithm}
\usepackage{algorithmic}
\usepackage{amssymb}
\usepackage{newfloat}
\usepackage{listings}
\usepackage[most]{tcolorbox}
\usepackage{xcolor}
\usepackage{enumitem}
\usepackage{caption}
\DeclareCaptionStyle{ruled}{labelfont=normalfont,labelsep=colon,strut=off} 
\floatstyle{ruled}
\newfloat{listing}{tb}{lst}{}
\floatname{listing}{Listing}

\title{Beyond I'm Sorry, I Can't: Dissecting Large-Language-Model Refusal}

\author{
    Nirmalendu Prakash\textsuperscript{\rm 1}\thanks{Correspondence to: nirmalendu\_prakash@mymail.sutd.edu.sg}\equalcontrib,
    Yeo Wei Jie\textsuperscript{\rm 2}\equalcontrib,
    Amir Abdullah\textsuperscript{\rm 3}\equalcontrib,
    Ranjan Satapathy\textsuperscript{\rm 4},
    Erik Cambria\textsuperscript{\rm 2},
    Roy Ka-Wei Lee\textsuperscript{\rm 1}
}
\affiliations{
    \textsuperscript{\rm 1}Singapore University of Technology and Design\\
    \textsuperscript{\rm 2}Nanyang Technological University\\
    \textsuperscript{\rm 3}Thoughtworks\\
    \textsuperscript{4}Institute of High Performance Computing (IHPC),\\ Agency for Science, Technology and Research (A$\textasteriskcentered$ STAR)
}

\begin{document}

\maketitle

\begin{abstract}
Refusal on harmful prompts is a key safety behaviour in instruction‑tuned large language models (LLMs), yet the internal causes of this behaviour remain poorly understood.
We study two public instruction tuned models—Gemma‑2-2B‑IT and LLaMA‑3.1-8B‑IT using sparse autoencoders (SAEs) trained on residual‑stream activations.
Given a harmful prompt, we search the SAE latent space for feature sets whose ablation flips the model from refusal to compliance, demonstrating causal influence and creating a jailbreak.
Our search proceeds in three stages: 1. Refusal Direction - Finding a refusal mediating direction and collecting SAE features close to that direction, followed by 2. Greedy Filtering - to prune this set to obtain a minimal set and finally 3. Interaction Discovery - a factorization‑machine (FM) model that captures non‑linear interactions among the remaining active features and the minimal set.  This pipeline yields a broad set of \textit{jailbreak-critical} features, offering insight into the mechanistic basis of refusal. Moreover, we also find evidence of redundant features which remain dormant unless earlier features are suppressed.
Our findings highlight the potential for fine-grained auditing and targeted intervention in safety behaviours by manipulating the interpretable latent space.
\end{abstract}

\begin{links}
    \link{Code}{https://github.com/Social-AI-Studio/LLM_refusal_interp}
    \link{Extended version}{https://arxiv.org/abs/2509.09708}
\end{links}

\input{sections/intro}
\input{sections/related_work}
\input{sections/experiment_setup}

\input{sections/methodology3}

\input{sections/methodology2}

\input{sections/results}

\input{sections/SAE_size_impact}
\input{sections/conclusion}
\input{sections/limitations}
\section{Acknowledgments}
This research/project is supported by the Ministry of Education, Singapore under its MOE Academic Research Fund Tier 2 (MOE-T2EP20123-0005). The research is also supported by the National Research Foundation, Singapore under its National Large Language Models Funding Initiative (AISG Award No: AISG-NMLP-2024-005). Any opinions, findings and conclusions or recommendations expressed in this material are those of the author(s) and do not reflect the views of the National Research Foundation and AI Singapore.
\bibliography{aaai2026}

\input{sections/appendix}
\end{document}

%% file: sections/intro.tex
\section{Introduction}
Refusal is a cornerstone safety behavior in aligned LLMs. When presented with potentially harmful, illegal, or unethical requests, an aligned model is expected to decline to respond, commonly through a refusal message such as ``I'm sorry, but I can't help with that.'' Yet public “jailbreak” leaderboards continue to show that adversarial prompts can bypass safety filters \cite{shen2025pandaguard}, while overly cautious models sometimes decline perfectly benign requests \cite{xie2024sorry, rottger2023xstest}. 
These twin failure modes underline the need to understand and adjust refusal behaviour mechanistically rather than relying on trial‑and‑error alignment.

A growing body of literature demonstrates that carefully crafted text, prefix or multi-shot prompts can bypass guard‑rails across model families \cite{wei2023jailbroken, anil2024many}. Once a public exploit emerges, safety teams must scramble to patch it, often tightening filters in ways that exacerbate over‑refusal. 

Mainstream defences center on behavioural training - supervised fine‑tuning (SFT) and Reinforcement Learning from Human Feedback (RLHF, RLAIF) - combined with post‑hoc policy filters~\cite{Ouyang2022Training,Bai2022Training,Bai2022Constitutional}. 
Although RL techniques improve harmlessness, it inherits classic RL failure modes such as reward hacking and brittleness under distribution shift~\cite{skalse2022defining}.  
Recent reward‑shaping schemes~\cite{rafailov2023direct} alleviate some issues, yet they remain iterative \emph{trial‑and‑error} tweaks to an opaque system, optimising outputs without revealing \emph{why} the model refuses or complies. Moreover, these methods require an auxiliary reward model and on-policy optimisation, making them impractical at inference time, whereas SAE or steering interventions add only a lightweight mask in a single forward pass.


Mechanistic interpretability seeks to reverse‑engineer neural networks into human‑understandable components.  
Sparse autoencoders (SAEs) have become a key tool in this agenda: by disentangling highly superposed vectors into a moderately over‑complete set of sparse latent features that often align with semantic concepts~\cite{bricken2023monosemanticity}.  
Recent work \cite{arditi2024refusal} shows that ablating a single residual-stream direction can flip refusals. \citet{wollschlager2025geometry} find that the refusal mechanism is governed by complex spatial structures. Another work by \citet{o2024steering} shows that steering or ablating few SAE features can flip refusal behavior on Phi-3 Mini. These findings suggest that refusal is mediated by a \emph{sparse} sub‑circuit rather than a diffuse behavioral mask.

At the same time, LLM computation exhibits substantial \emph{redundancy}: multiple, partially overlapping features can implement the same logical function, a phenomenon dubbed the \emph{hydra effect}~\cite{mcgrath2023hydra}.  
When one head (feature) is severed by ablation, another may activate to preserve the downstream behaviour.  
In our experiments we observe precisely this pattern: features we identify as critical to refusal re‑activate whenever a distinct precursor is removed, implying a safety‑critical network of “hydra‑heads’’ whose dependencies must be mapped \emph{jointly}. 
Understanding and managing this redundancy is essential for reliable, feature‑level alignment interventions. However, the standard tool of linear probes implicitly models effects as weighted sums of individual features - i.e., they assume each feature contributes independently and linearly to the output. As a result they cannot capture higher-order interactions, correlations, conditional dependencies, or shared variance arising from redundancy \citep{nonlinear_interactions_0, nonlinear_interactions_1, nonlinear_interactions_2}.

We address these gaps by tracing refusal back to causal SAE features, and identifying feature correlations more carefully. Our contributions can be summarized as follows:

\begin{enumerate}
\item The first end‑to‑end pipeline for locating causal refusal features via SAEs and targeted ablation.
\item Introducing a factorization machine based approach to recover related features.
\item Discovery of redundancy in LLM refusal mechanism.
\item Open‑sourced code and feature indices to accelerate future work on safe and interpretable LLMs.
\end{enumerate}



%% file: sections/related_work.tex
\section{Related Works}
\begin{figure*}[t]  
    \centering
    \includegraphics[width=1\textwidth]{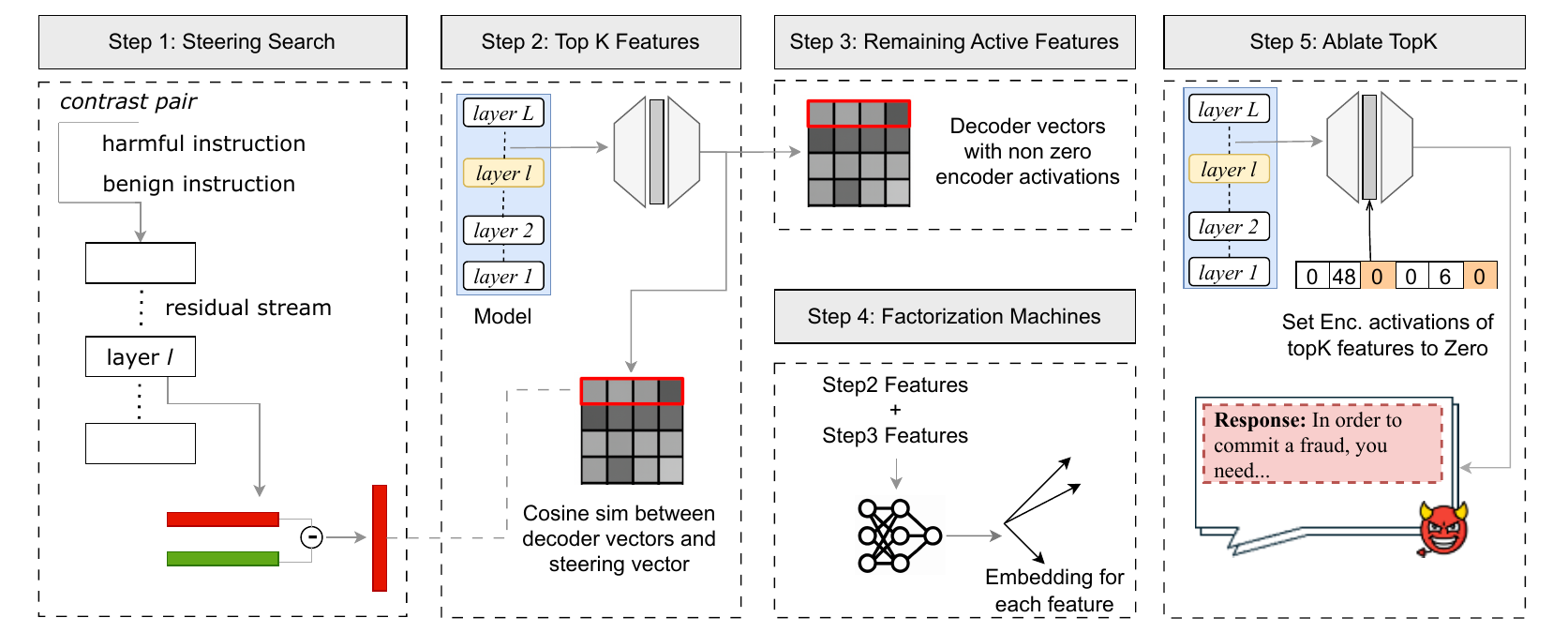}
    \caption{Figure shows the steps involved in causal feature search. Trapezium shapes represent encoder and decoder of a SAE, with the rectangular block in between representing the sparse encoder activations, where ablation is applied.}
    \label{fig:steps}
\end{figure*}
\subsection{LLM Alignment}
The goal of aligning large language models (LLMs) is to ensure that outputs remain helpful, harmless, and consistent with human preferences. A central behaviour shaped by alignment training typically via reinforcement learning from human feedback (RLHF) is \textit{refusal}, in which the model declines to answer unsafe or harmful prompts. This is often implemented via fine-tuning or reward models that penalize completions violating safety guidelines. Several studies analyze refusal rates across domains \citep{ganguli2023capacity}, examine adversarial vulnerabilities \citep{zou2023universal}, or develop red-teaming methods to test refusal robustness \citep{perez2022red}. However, by focusing solely on refusal as an observable outcome,  these approaches overlook the fundamental question of which internal reasoning pathways actually produce this behavior.

\subsection{Mechanistic Interpretability of Alignment}
Mechanistic interpretability aims to open the black box of neural networks by tracing how activations and weights implement behavior \citep{olah2020zoom, elhage2021mathematical}. In alignment research, this includes identifying circuits responsible for factual recall \citep{chughtai2024summing}, chain-of-thought reasoning \citep{dutta2024think}, or finding safety-mediating neurons \citep{chen2024finding}. Prior work has shown that refusals can sometimes be controlled by manipulating single activation directions \citep{arditi2024refusal}, and that harmful and harmless prompts cluster distinctly in activation space \citep{jain2024makes}. \citet{lindsey2025biology} investigate the internal mechanisms behind model refusals on a single jailbreak prompt, but do not arrive at a definitive explanation. A recent work by \citet{o2024steering} proposes clamping SAE features at inference time to steer refusal behavior, but also cautions that this intervention can degrade the model’s overall performance. These results suggest that refusal is driven by coherent internal structures, which compel further analysis into what these structures are.



\subsection{Sparse Autoencoders}
A sparse autoencoder (SAE) is a neural network trained to compress and reconstruct activations while encouraging most latent units to remain inactive for any given input. This sparsity constraint forces the model to learn a set of meaningful, disentangled features that capture distinct patterns in the data \cite{elhage2021mathematical,sae-proposition,bricken2023monosemanticity,templeton2024scaling}. The sparse latents are more interpretable and monosemantic than directions identified by alternative approaches \citep{cunningham2023sparse}. Neuronpedia\footnote{\url{https://www.neuronpedia.org}} offers an actionable vocabulary interpreting these directions.

Given a hidden representation $h \in \mathbb{R}^d$, its SAE reconstruction is given by $\hat{h} = W_\text{dec} f(W_\text{enc} h)$ , where $W_\text{enc}$ and $W_\text{dec}$ are the encoder and decoder matrices, and $f$ is a sparse activation function\footnote{Bias terms are omitted for brevity}. For a set of target features $S \subset \{1, \ldots, k\}$, we \textit{zero-ablate} their contribution by replacing the reconstructed vector with:
\[
\tilde{h}
=
W_{\text{dec}}\!\bigl(f(W_{\text{enc}}h)\odot\mathbf 1{S'}\bigr)
\;+\;
\bigl(h - W_{\text{dec}}f(W_{\text{enc}}h)\bigr)
\tag{1}
\]
where each dimension of the sparse activation $z = f(W_{\text{enc}} h) \in \mathbb{R}^k$ has a corresponding row in $W_\text{dec}$ that contributes additively to the reconstruction and interpreted as a feature. $\mathbf{1}{S'}$ is a binary mask that zeros out all features in $S$. This allows us to isolate the functional role of a sparse set of latent dimensions in driving model refusal.

%% file: sections/experiment_setup.tex
\section{Experimental Setup}

\begin{figure*}[t]  
    \centering
    \includegraphics[width=0.95\textwidth]{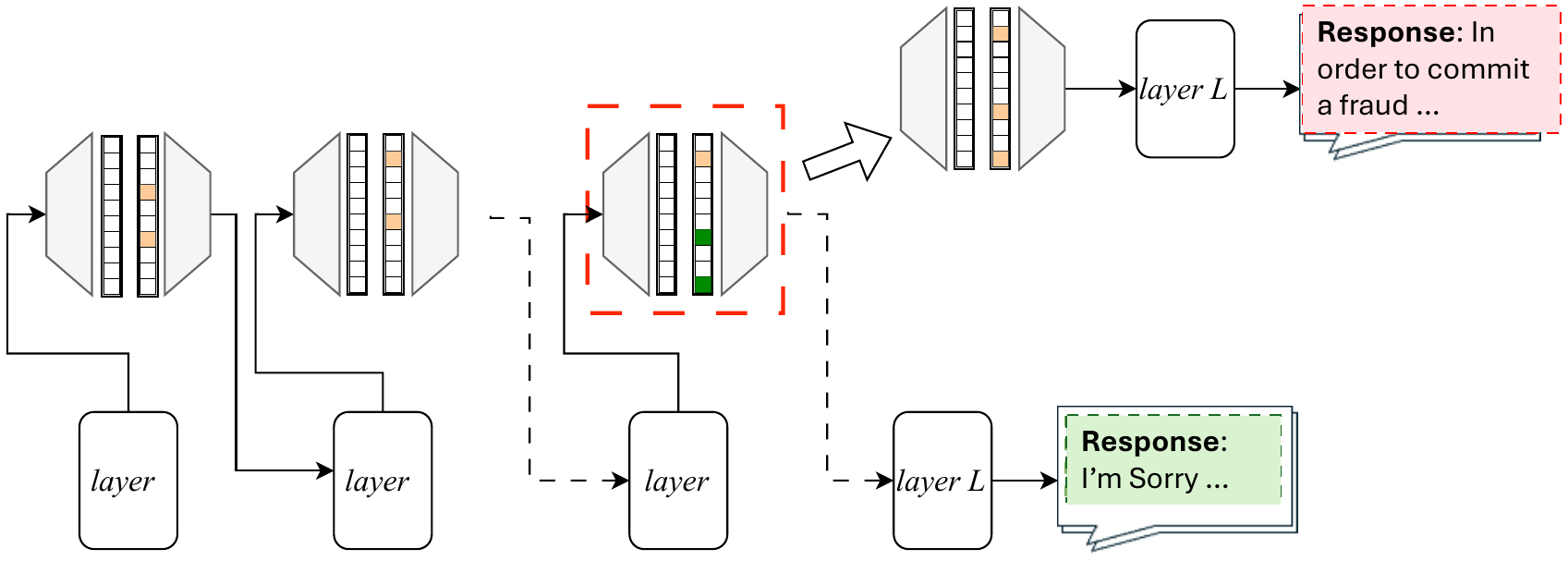}
    \caption{Shown above is part of the computation flow in a decoder only LLM. Attached to a layer is a SAE. Square boxes denote SAE encoder activations. Orange denotes ablated feature and green activated. Here, we demonstrate that ablating some of the early layer features (orange) can activate (green) a set of features in a downstream layer. These downstream features are causal to refusal in-spite of being not active in the first place. Shown on top right is the response (jailbroken) after ablating these (orange+green) and on the bottom right is the safe response when these are not ablated.}
    \label{fig:redundant_features}
\end{figure*}

\subsection{LLMs}
We conduct our study on Gemma2--2B-Instruct and LLaMA-3.1-8B-Instruct. Hereafter, we refer to the two models as \textsc{LLaMA} and \textsc{Gemma} in the paper for brevity. The SAEs (Gemmascope and Llamascope) are available in a range of sizes and sparsity settings. The settings used for the experiments are available in Appendix section ``Experimental Setup''.

\subsection{Dataset}
\label{datasets}
\begin{enumerate}
    \item Analysis: To identify the SAE latents that mediate refusal, we use the $D_{\text{harmful}}$ dataset curated by \citet{arditi2024refusal}, which aggregates harmful prompts from several widely used safety datasets. We verify and filter the samples in $D_{\text{harmful}}$ that do not elicit refusal on our models tested. 


\item Unsafe Taxonomy - Coconot: \citet{brahman2024art} introduce a taxonomy of non-compliance, outlining when and how language models should refuse to comply with user instructions. This taxonomy includes five categories: \textit{incomplete}, \textit{unsupported}, \textit{indeterminate}, \textit{humanizing}, and \textit{unsafe} prompts. In our analysis, we focus on the \textit{unsafe} prompts subset. Safety violations are further subdivided into five categories: ``Triggers for Offensive Language'', ``Dangerous or Sensitive Topics'', ``False Information'', ``Privacy Violations'', and ``Copyright Violations'', comprising a total of 2,586 samples.

\end{enumerate}

\subsection{SAE Feature Explanation through Neuronpedia}
To interpret the discovered SAE features, we obtain natural language explanations from Neuronpedia, which uses a GPT-4o-mini model to generate descriptions based on top activating samples. These samples are drawn from a large-scale dataset, which spans a wide range of text domains. We retrieve explanations using the official Neuronpedia API\footnote{\url{https://www.neuronpedia.org/api/feature/{modelId}/{layer}/{index}}}
, by querying each feature with its model, layer, and index identifiers.

\subsection{Safety Metric}
 \citet{arditi2024refusal} match against a set of phrases such as `I'm sorry', `As an AI' etc. to measure refusal. To asses whether the completions contain harmful content, authors also use \textsc{LLaMA GUARD 2} \cite{metallamaguard2}. We find that degenerate responses classified as safe by \textsc{LLaMA GUARD 2} do not represent jailbreak. Also, string matching is limited by the set. For this reason we use HarmBench Classifier model \cite{mazeika2024harmbench} to accurately determine harmful responses and hence Jailbreak. 

\subsection{Attack Success Rate (ASR)}
We report Attack Success Rate (ASR) as the fraction of harmful prompts that elicit non-refusal responses from the model. We measure ASR on $D_{\text{harmful}}$. When reporting on subsets of the data, we mention ``Jailbreak success'', so as not to confuse with the overall values.

%% file: sections/methodology3.tex
\section{Experiments}
The observation that ablating a single steering direction across layers
is sufficient to jailbreak several aligned models suggests that this
direction carries safety‑critical information throughout the network.
Furthermore, \citet{jain2024makes} show that hidden states for safe and
unsafe prompts begin to diverge from an early layer.
These findings motivate the following simple diagnostic:
for each layer we compute the cosine similarity between the residual‑stream
activation and a \textit{refusal mediating} direction, using \citet{arditi2024refusal}'s approach.

Applied to \textsc{Gemma} and \textsc{LLaMA},
the diagnostic reveals two consistent patterns
(see Appendix Section ``Preliminary Analysis''):
\begin{enumerate}
    \item Harmful‑prompt activations show markedly higher similarity to the
          steering vector than benign‑prompt activations.
    \item This similarity grows monotonically up to the layer at which the
          steering vector is measured.
\end{enumerate}
The early‑layer alignment of harmful activations with the \emph{refusal}
direction indicates that refusal is mediated by a sparse sub‑circuit that is
amplified through the forward pass.

Guided by this insight, we propose a three‑stage causal‑feature discovery
pipeline:
\begin{enumerate}
  \item Refusal mediating direction.
        Obtain an effective refusal steering vector and select the top‑$K$ SAE
        features whose decoders are most strongly aligned with it.
  \item Greedy filtering.
        Iteratively ablate features to obtain the minimal subset whose removal
        flips the model from refusal to compliance, thereby establishing direct
        causality.
  \item Interaction discovery.
        Feed the remaining active features and the minimal causal set into a
        factorization machine to uncover additional features.
\end{enumerate}

We detail the stages below:
\subsection{Stage 1: Refusal Mediating Features}
\begin{algorithm}[t]
\caption{Block-Wise Selection of a Faithful Latent Set}
\label{alg:block}
\begin{algorithmic}[1]
\STATE \textbf{Input:} Prompt $p$; candidate pool $C$; block size $N$; threshold $\tau$
\STATE \textbf{Output:} Faithful latent set $K$
\STATE $K \gets \emptyset$
\WHILE{$|C \setminus K| \ge N$}
    \STATE Draw a random block $V \subseteq C \setminus K$ with $|V| = N$
    \IF{$\Delta(p, K, V) \ge \tau$}
        \STATE $K \gets K \cup V$
    \ELSE
        \STATE $C \gets C \setminus V$
    \ENDIF
\ENDWHILE
\STATE \textbf{return} $K$
\end{algorithmic}
\end{algorithm}
Our goal is to identify, for every harmful prompt \(s\), a \emph{minimal} set of SAE features \(M_s\) whose ablation is sufficient to jailbreak the response.
The procedure consists of two steps.

\paragraph{1.~Deriving a refusal steering vector.}
Following \citet{arditi2024refusal}, we compute a
\textit{steering vector}~\(\mathbf v\) for each model by taking the
difference of residual-stream activations between (i) prompts that elicit a
refusal and (ii) matched benign prompts.  Normalizing~\(\mathbf v\) yields a
unit direction that consistently nudges the model towards compliance when
subtracted from the residual stream. 
For details on how this direction is derived, see section ``Deriving a Refusal Steering Vector'' in Appendix. Sample jailbreak responses are available in section "sample Jailbreaks" in Appendix.

\paragraph{2.~Selecting a Top-\(K\) candidate set.}
For every SAE latent \(z\) with decoder weight
\(\mathbf d_z\in\mathbb R^{d_{\text{model}}}\) we compute the
cosine similarity \(\cos(\mathbf v,\mathbf d_z)\).
We create a list of latents by taking the
top \(K\) in each layer as our initial candidate set
\(\mathcal L_K\!=\!\{z_1,\dots,z_K\}\).
We start with \(K{=}10\) and increase in steps of 10 until ablating
\(\mathcal L_K\) successfully jailbreaks the prompt
(validated with the refusal classifier HarmBench \citep{mazeika2024harmbench}).
For Gemma we limit our analysis to \(K\!=\!200\), whereas for
LLaMA a single step (\(K\!=\!10\)) results in degenerate responses. we find that Layers 1 and 2 typically encode grammatical features and ablating them results in degenerate responses. Thus we exclude these features corresponding to these layers in stage 1. 

\subsection{Stage 2: Greedy Pruning To A Minimal Faithful Set}
\citet{arditi2024refusal} observe that the probability assigned to the first‐person pronoun ``\texttt{I}'' at the final prompt position tracks jailbreak success remarkably well. Using this finding, starting from \(\mathcal L_K\) we iteratively remove features that do not cause a substantial change to ``\texttt{I}'' token logit at output.  Algorithm~\ref{alg:block} describes the pruning procedure, which is a simplified variant of the incompleteness algorithm from \citet{wang2022interpretability}. 


Let \(C\) be the full top-$K$ candidate pool of SAE features and
\(K=\varnothing\) the set of features already accepted.
At each iteration we sample a block  
\(V\subseteq C\setminus K\) of fixed size \(N\).
Define
\[
P\bigl(p,\mathcal A\bigr)
     \;=\;
\Pr\!\bigl(\texttt{I}\mid p,\text{ model ablated on }\mathcal A\bigr),
\]
the probability of emitting the token “\texttt{I}’’ for prompt \(p\)
after ablating latent set \(\mathcal A\).
The relative impact of ablating the block \(V\) is then
\begin{equation}
\label{eq:blockfaith}
\begin{gathered}
\Delta(p,K,V) = \\[3pt]
\displaystyle
\frac{\Bigl|\,
      P\!\bigl(p,\,C\setminus K\bigr)\;-\;
      P\!\bigl(p,\,C\setminus(K\cup V)\bigr)
      \Bigr|}
     {P\!\bigl(p,\,C\setminus K\bigr)} .
\end{gathered}
\end{equation}

A higher \(\Delta\) implies a larger change in the “\texttt{I}’’
probability and therefore greater faithfulness of the candidate block.

If \(\Delta(p,K,V)\ge\tau\) for a threshold \(\tau\), the subset is deemed
\emph{faithful} and we update \(K \leftarrow K\cup V\); otherwise \(V\) is
discarded. We fix
\(N{=}5\) and sweep the threshold
from 0.1 to 0.8 in increments of 0.1. Also, for each threshold, we run the greedy pruning algorithm for three random seeds and keep only those features which are observed with all seeds. 

\paragraph{Results.}
\textsc{Gemma}’s most effective steering vector emerges at layer~16, whereas for \textsc{LLaMA} it appears at layer~13. 



\begin{table}[t]
\centering
\begin{tabular}{lcc}
\hline
                                   & \textbf{Gemma} & \textbf{LLaMA} \\ \hline
$D_{\text{harmful}}$               &   861             & 861               \\
ASR (no ablation)            &     4           &   71             \\
ASR (after ablation)         &    0.33            &   0.70             \\ 
Unique SAE Features                &     2538           &   110             \\ \hline
\end{tabular}
\caption{Jailbreak statistics for \textsc{Gemma} and \textsc{LLaMA} using features obtained after Stage 2.}
\label{tab:stage2_results}
\end{table}

After applying the pruning algorithm, we obtain 110 unique features for \textsc{LLaMA} and 2,538 features for \textsc{Gemma}, as summarized in Table~\ref{tab:stage2_results}. To semantically interpret the range of concepts encoded by these features, we retrieve their explanations from Neuronpedia and manually group them into thematic categories. 

We find that a subset of features encode concepts related to \textit{harm} and \textit{violence}, while a majority, across both models are associated with \textit{programming constructs} and \textit{punctuation}. These frequently occur across the studied layers. We hypothesize that this pattern may arise either due to limitations of the dataset used for generating the explanations or because these features encode general grammatical structure. Further analysis and grouped examples are presented in Appendix Section ``Feature-semantics''.

%% file: sections/methodology2.tex
\subsection{Stage 3 : Interaction Discovery}
As a first sanity check we asked a seemingly simple question:
Do the jailbreak–critical features returned by Stage~2 actually fire on the very prompts for which they are deemed causal?
For every harmful prompt we inspected the raw SAE activations and discovered—
surprisingly—that several (77 for LLaMA and 1656 for Gemma) ``critical’’ features were \textit{inactive} (zero activation).  Removing these inert features and re‑running the ablation caused the
jailbreak to fail (for LLaMA jailbreak success drops to 372 samples and for Gemma 103 samples), demonstrating that apparently silent units can still be
necessary. The only plausible explanation is that ablating some active features
allows previously silent features to switch on and compensate. This is direct evidence of the \textit{non‑linear hydra effect} among jailbreak–critical features; Figure \ref{fig:redundant_features} shows this effect. We study this phenomenon in more detail in the next section.

Having established that a minimal causal set can recruit additional units
through interaction, we next map the remainder of the causal neighborhood.
To capture higher‑order dependencies we fit a second‑order
Factorization Machine (FM) to the SAE activations, using a
combined corpus of \(D_{\text{harmful}}\) prompts and an equal number of benign samples from
\textsc{Alpaca} (dataset details in Appendix Section ``Experimental Setup''). Below, we briefly describe the FM formulation.


\begin{table}[t]
\centering
\begin{tabular}{lcc}
\hline
                                   & \textbf{Gemma} & \textbf{LLaMA} \\ \hline
$D_{\text{harmful}}$               &   861             & 861               \\
ASR (no ablation)            &   4             &   71             \\
ASR (after ablation)         &   0.31            &   0.57             \\ 
Unique SAE Features                &   2702             &   556             \\ \hline
\end{tabular}
\caption{Jailbreak statistics for \textsc{Gemma} and \textsc{LLaMA} using features obtained after Stage 3.}
\label{tab:stage3_results}
\end{table}

\paragraph{Factorization Machines.}
Factorization Machines (FMs) \cite{rendle2010factorization} are supervised learning models that unify the advantages of linear regression and matrix factorization. They are particularly effective in sparse, high-dimensional settings such as recommender systems. FMs model not only the linear effects of individual features but also capture pairwise feature interactions using low-dimensional latent vectors.

For a feature vector $\mathbf{x} \in \mathbb{R}^d$, the prediction of a 2-way factorization machine is given by:

\begin{equation}
\hat{y}(\mathbf{x}) = w_0 + \sum_{i=1}^{d} w_i x_i + \sum_{i=1}^{d} \sum_{j=i+1}^{d} \langle \mathbf{v}_i, \mathbf{v}_j \rangle x_i x_j
\end{equation}

Here, $w_0 \in \mathbb{R}$ is the global bias, $\mathbf{w} \in \mathbb{R}^d$ is the vector of weights for individual features, and $\mathbf{v}_i \in \mathbb{R}^k$ is the $i$-th latent embedding vector representing feature $i$, where $k$ is the dimensionality of the latent space. The dot product $\langle \mathbf{v}_i, \mathbf{v}_j \rangle$ captures the interaction between features $i$ and $j$, allowing the model to learn pairwise non-linear interactions efficiently.

This formulation allows FMs to generalize matrix factorization while also incorporating side information and high-order interactions with linear complexity.

We experiment with FM embedding sizes of 5, 20, and 50. An embedding size of 5 yields overly similar feature representations (cosine similarity $> 0.9$), indicating poor expressiveness. Increasing to 20 provides more diverse and meaningful embeddings. While size 50 offers marginal improvements, it incurs significantly higher training cost. We therefore use an embedding size of 20 in all subsequent experiments.

\paragraph{Results.}
We leverage feature activations at special tokens following the user instruction, as these have been shown to play a key role in driving refusal behavior~\cite{zhao2025llms}. To identify meaningful additions to our feature set, we rank the newly discovered features by computing the cosine similarity between their Factorization Machine (FM) embeddings and those of the Stage 2 features. We then select the top-$K$ features for ablation, sweeping $K$ from 100 to 2000 in increments of 100.

For \textsc{LLaMA}, we observe that beyond $K = 2000$, model outputs begin to degrade into degenerate or uninformative responses. In contrast, \textsc{Gemma} continues to produce jailbreak responses even beyond $K = 2000$. However, due to computational constraints, we limit our experiments to $K = 2000$. The results are summarized in Table~\ref{tab:stage3_results}. We note that a finer-grained search over the similarity threshold may uncover additional jailbreakable samples.

The observed drop in Attack Success Rate (ASR) compared to Stage 2 can be attributed to the removal of \textit{redundant} features, which, while overlapping, appear to contribute non-trivially to the success of jailbreaks.

We also analyze the semantic content of the newly added features using Neuronpedia explanations. In \textsc{LLaMA}, Stage 3 introduces several features associated with \textit{legal} or \textit{regulatory} content, while a majority represent highly specific, non-harm-related concepts, such as references to \textit{financial information}. A similar trend is observed in \textsc{Gemma}, where Stage 3 surfaces more fine-grained and concrete feature representations. For a more detailed breakdown, refer to Appendix Section ``Feature Semantics''.

%% file: sections/results.tex
\section{Analysis}
We refer to features available after stage 2 as ``Stage 2'' features and features added in stage 3 as ``Stage 3'' features. Next, we examine how the features identified in Stage 3 influence the Stage 2 features.

\paragraph{Causal link between Stage~2 and Stage~3 features}
We test two questions: (i) Are Stage~3 features alone sufficient to jailbreak (i.e., independently causal for refusal)? (ii) How strongly do Stage~3 and Stage~2 features causally influence each other?
On \textsc{LLaMA}, ablating only Stage~3 features jailbreaks $330/372$ (\(89\%\)) of cases; on \textsc{Gemma}, only \(8\) (\(3\%\)). 
For Stage~3\(\rightarrow\)Stage~2 on \textsc{LLaMA}, ablating Stage~3 deactivates on average \(12\%\) (up to \(44\%\)) of active Stage~2 features, and reduces activation by \(81\%\) in the remainder, indicating a strong dependency; on \textsc{Gemma}, it deactivates only \(3\%\) with no activation drop, suggesting a more diffuse mechanism.
Conversely, for Stage~2\(\rightarrow\)Stage~3, \textsc{Gemma} shows \(100\%\) deactivation, while \textsc{LLaMA} shows \(10\%\) deactivation and a \(32\%\) mean activation drop in Stage~3 features.

We next compare Stage~3 features identified by FMs versus a linear probe.

\paragraph{Factorization Machines vs. Linear Probe}
To verify the existence of non-linear feature interaction, we compare with a linear probe trained on the same feature activations.
This time, we rank features by the magnitude of their learned weights in a similar fashion and pick top-$K$ features to consider for ablation. We do this for \textsc{LLaMA}, using $K \in \{100, 500, 1000, 2000\}$ and evaluate their ability to trigger jailbreaks when ablated on \textsc{$D_{harmful}$} samples.

We find that these top-$K$ features lead to jailbreak in significantly fewer cases (only 101 samples) compared to the FM-based approach. This suggests that refusal-inducing features interact in a fundamentally non-linear manner. Next, we investigate the redundant features we discussed above.

\paragraph{Redundant Features}
To analyze the redundant features, we follow these steps:
\begin{enumerate}
    \item From the features found in Stage 2, for each of the samples, we find which of the jailbreak critical features do not fire.
    \item Next, we ablate just the non-active set and record the impact on jailbreak. If there is no impact, we discard such samples from further analysis.
    \item For the remaining set of samples, we ablate the active feature set and record activations on non-active set, along with the tokens on which they activate.
\end{enumerate}

On \textsc{LLaMA}, we find that out of the redundant features, \textbf{74\%} become active on system tokens after ablating the active set. Of these, approximately \textbf{97\%} fire on the \texttt{<|begin\_of\_text|>} token. On \textsc{Gemma}, we find that the \textit{redundant} features fire on a range of tokens but still \texttt{<bos>} token records highest activity (\textbf{16\%}) for these features.

To further investigate this, we obtain explanations for these features from  Neuronpedia and find that about 31\% contain text related to punctuation and programming syntax on \textsc{LLaMA}. On \textsc{Gemma} as well we find a number of feature explanations about programming and punctuation. We also find some features related to harm and uncertainty.

To probe their causal role further, we steer the models by clamping the above features to a higher positive value, on the \texttt{<bos>} token for Gemma and \texttt{<|begin\_of\_text|>} token for \textsc{LLaMA}, and presenting the model with otherwise benign requests such as ``Tell me a story'' or ``What are the benefits of meditation?''. We find that on \textsc{Gemma}, the model responses indicate harm (e.g. ``Meditation is a serious and potentially harmful practice.'') for smaller values and refusal for larger values (e.g. ``Meditation is illegal and I will not answer your question.''). We do not observe harmful responses on  \textsc{LLaMA} but we do observe instances of token ``no'' (e.g. ``Meditation is a meditation is no no'') which may indicate a tendency towards refusal. Example generations are provided in Appendix Section~`Redundant Features'; a fuller exploration of this phenomenon is left to future work.


\begin{figure*}[t]  
    \centering
    \includegraphics[width=1\textwidth]{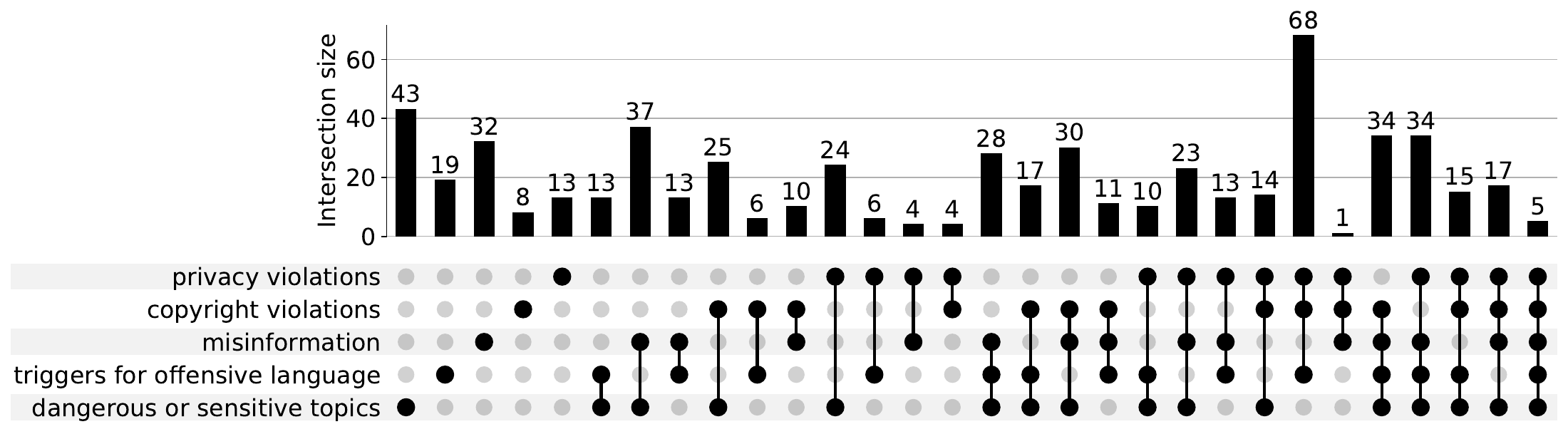}
    \caption{The figure shows harm types based on the \textit{Coconot} unsafe taxonomy and the count of feature activations against each type (first four show individual types and remaining on the right show count of features which fire on multiple harm types.) }
    \label{fig:harm_taxonomy_mapping}
\end{figure*}

\paragraph{Comparison with random features}
To assess the significance of the jailbreak-critical features identified, we perform a control experiment using an equal number of randomly selected active features for each prompt. We then compare jailbreak performance between the two settings. 
Ablating these random features results in no jailbreak across samples for \textsc{LLaMA} and 5 samples for \textsc{Gemma}. This highlights that the jailbreak-critical features play a meaningful and non-random role in refusal behavior. 

\paragraph{Model performance loss by ablating jailbreak critical features}
To ascertain the downstream impact of ablating these features on LLMs’ cross-entropy (CE) loss
(see Appendix Section “Downstream Performance”), we compare CE with and without feature ablation.
The CE loss is slightly higher with feature ablation compared to direction ablation, as features
represent a variety of directions and not just refusal.




\paragraph{Causal feature finding using alternate methods}
\citet{o2024steering} find refusal mediating features on Phi-3 Mini model by selecting features activating on at least two tokens on a harmful prompt. This is further filtered using a clamping hyperparameter search where the features are clamped to a range of positive values. We adapt the first part of their approach to do a hyperparameter search over different Ks (features activating on at least k tokens). We do this over each sample and ablate the features found. We obtain a maximum ASR of 0.1 on \textsc{Gemma} and 0.07 on \textsc{LLaMA}. Refer to Appendix Figure 1 for ASR across different k.


Next, we map the features found to a ``safety specific non-compliance" taxonomy.


\paragraph{Mapping causal features to unsafe taxonomy}

We map each features found on \textsc{LLaMA} to one or more of the harm categories on Coconot dataset by identifying non-zero activations on samples belonging to each category. The results of this mapping are shown in Figure~\ref{fig:harm_taxonomy_mapping}. We find that a majority of the features activate on ``Dangerous or Sensitive Topics'', often in conjunction with one or more additional harm types.

To validate these mappings, we conduct a human annotation study comparing activation-based labels with human interpretations of the features, derived from top activating tokens and Neuronpedia explanations. We sample an equal number of features that activate on a single harm category and those that activate on multiple categories, along with a matched number of randomly selected features. Annotators are shown the feature explanation and top activating samples, and are asked to select all applicable harm categories or choose ``None''. Annotation details are provided in Appendix Section ``Human Study''.

We employ three annotators and assign a final label based on majority agreement (i.e., at least two annotators concur). When compared to activation-based labels, we observe an accuracy of 13.42\%. We hypothesize that increasing the dimensionality of the SAE may yield more fine-grained and disentangled features, thereby improving alignment with human intuition.



We hope that such mappings can inform more fine-grained steering of model behavior, and that similar mappings can be constructed for other models as well. A more comprehensive investigation along these lines is left to future work.

%% file: sections/conclusion.tex
\section{Conclusion \& Future Work}



We present the first pipeline that combines SAEs with Factorization Machines to isolate causal features governing LLM refusal.  
FMs capture non-linear feature interactions better than linear probes, enabling more precise ablation. Our analysis also revealed \emph{redundant} “hydra” features that remain dormant until others are suppressed.  
A human study maps these latent features to an established unsafe-prompt taxonomy, laying groundwork for feature-level steering of refusal behaviour.






In future work, we plan to (i) characterize the safety benefits and risks of redundancy, (ii) evaluate how varied jailbreak strategies—role-play, multi-turn coercion, etc.—perturb the activation space (iii) analyze prompts that consistently resist jailbreak to uncover protective circuits and (iv) harden LLM refusal for specific reasoning types (e.g., legal reasoning).
These directions aim to advance transparent and controllable safety mechanisms for next-generation language models.

We hope that our findings inspire deeper mechanistic investigations into LLM refusal and lead to principled audits of LLM safety, enabling answers to questions such as: ``To what extent is refusal driven by legal reasoning?'' or ``How much do moral heuristics contribute to refusal decisions?''.


%% file: sections/limitations.tex
\section*{Limitations}
\begin{itemize}
    \item \textbf{Generality.} Our analysis focuses on two model sizes; larger models remain to be tested.
    \item \textbf{SAE stability.} SAEs are sensitive to data and initialization; \citet{paulo2025sparse} report only $\sim$30\% latent overlap across seeds, motivating replication across multiple SAE runs.

    \item \textbf{Computational cost.} Searching for jailbreak-critical latents and re-evaluating them across many benchmarks requires thousands of forward passes; and we plan to explore a matching-pursuit algorithm in future work.
\end{itemize}

%% file: sections/appendix.tex
\section{Experimental Setup}
\paragraph{SAE settings.} The \textsc{Llamascope} SAEs are available in 32k and 128k encoder sizes. SAEs are trained on the residual stream, MLP activations, and attention head outputs. In this work, we use the \textbf{residual stream} SAEs.
These SAEs are trained on Top-K design and also include JumpReLU for post processing.

The \textsc{Gemmascope} SAEs are available in 16k and 65k variants, trained using JumpReLU sparsity with varying $L_0$ values per layer. The $L_0$ setting in \textsc{Gemmascope} SAEs controls the number of non-zero (active) features allowed per activation vector in each layer, effectively limiting the sparsity level of the learned representation. The L0 settings used for \textsc{Gemmascope} are listed in Table \ref{tab:gemma65k_l0}.
\begin{table}[h]
    \centering
    \small
    \begin{tabular}{c|cccccccc}
    \hline
    \textbf{Layer} & 0 & 1 & 2 & 3 & 4 & 5 & 6 & 7 \\
     \hline
    $L_0$ & 43 & 54 & 77 & 42 & 46 & 53 & 56 & 57 \\
     \hline
    \textbf{Layer} & 8 & 9 & 10 & 11 & 12 & 13 & 14 & 15 \\
     \hline
    $L_0$ & 59 & 61 & 66 & 70 & 72 & 75 & 73 & 68 \\
     \hline
    \end{tabular}
    \caption{$L_0$ sparsity settings used for 65k Gemma SAE}
    \label{tab:gemma65k_l0}
    \end{table}

We used NVIDIA A100 80GB GPU for our experiments. We download the pretrained LLMs from Huggingface and use data format bfloat16 due to resource constraints. Also, we use SAELens package to work with \textsc{LLamascope} SAEs. 

\paragraph{Dataset for training Factorization Machines.} We obtain advbench set (521 samples) from Dharmful and an equal number of random samples from \textsc{alpaca}. We split the dataset into train-validation-test sets using proportion 80-10-10. Next, we record activations on the last 4 tokens, on layer 2-steering layer (layer corresponding to where the steering vector is obtained) on each of the samples. \textsc{Alpaca} samples are labeled 0 and advbench samples have labels 1. We train FM with a learning rate of 1e-3 for 20 epochs. Best performing epoch on validation set is saved and finally evaluated on the test set. We observe accuracy of 98\% on the test set for both the models (\textsc{LlaMA} and \textsc{Gemma}).
    
\section{Preliminary Analysis}
\begin{figure}[t]
    \centering
    \begin{minipage}{0.45\textwidth}
        \centering
        \includegraphics[width=\linewidth]{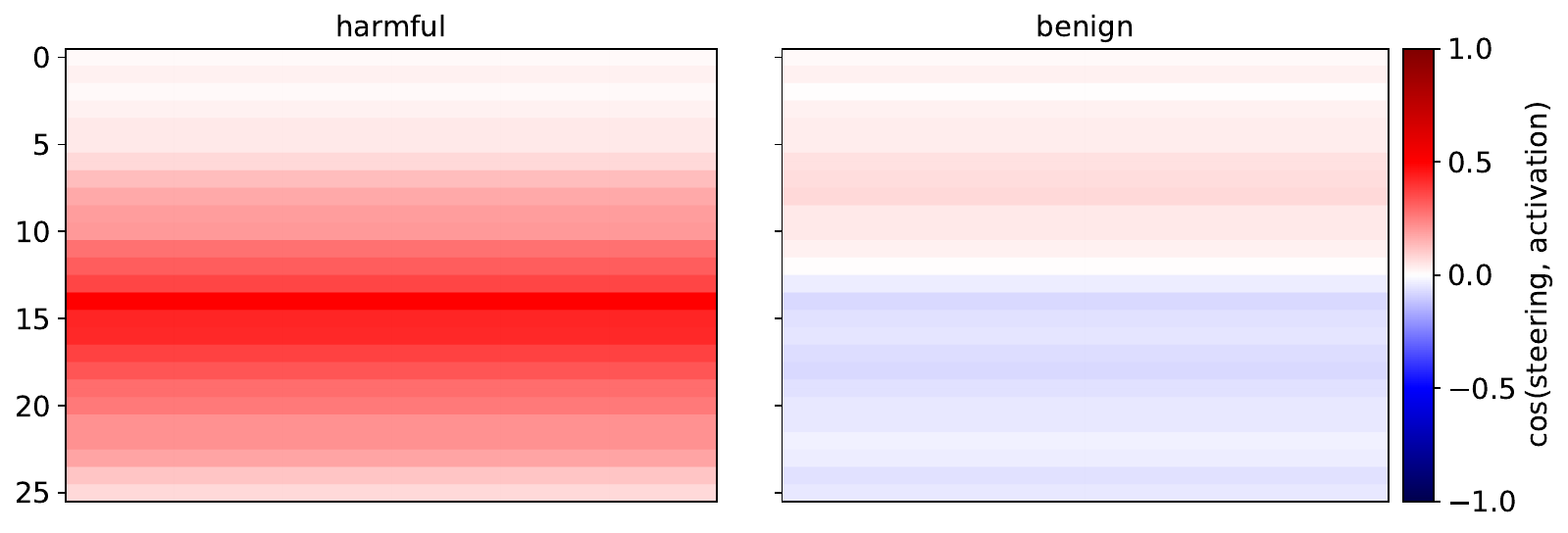}
        \caption{Figure shows cosine similarities across layers between residual activations and steering vector at the last token, for \textsc{Gemma}}
        \label{fig:prelim_study_gemma}
    \end{minipage}
    
    \begin{minipage}{0.45\textwidth}
        \centering
        \includegraphics[width=\linewidth]{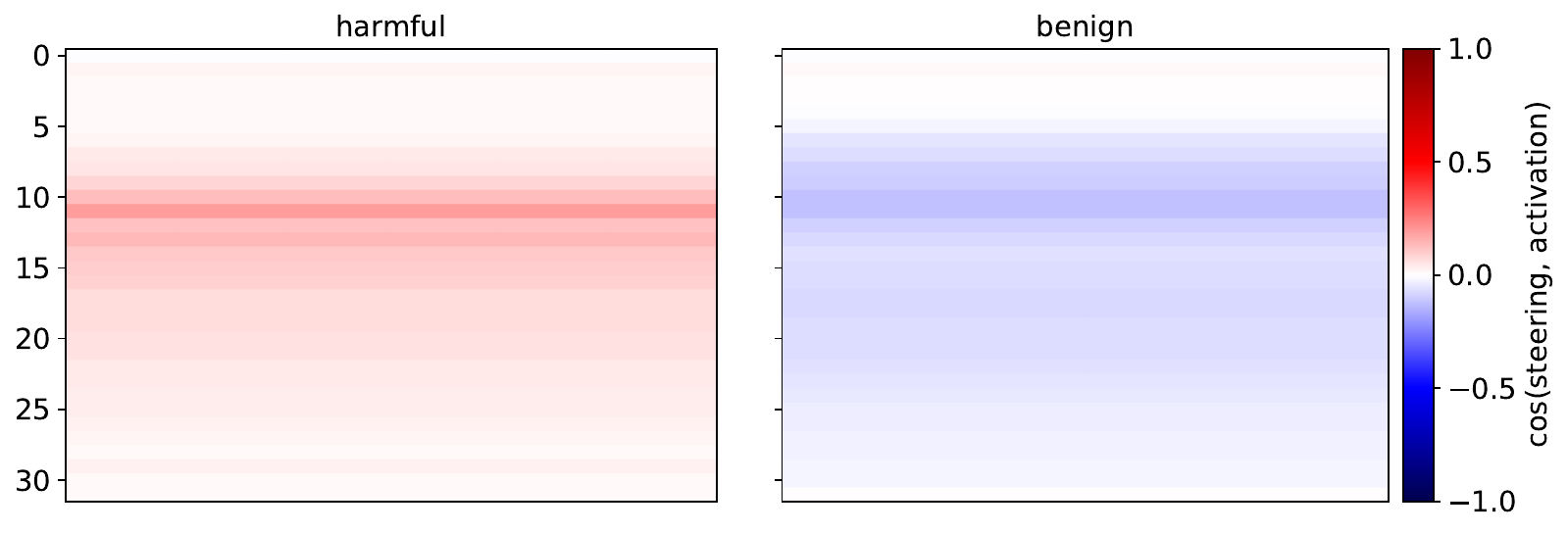}
        \caption{Figure shows cosine similarities across layers between residual activations and steering vector at the last token, for \textsc{LLaMA}}
        \label{fig:prelim_study_llama}
    \end{minipage}
\end{figure}

We measure cosine similarity of steering vector obtained on \textsc{Gemma} with the residual stream activations on the last token. Shown in Figure \ref{fig:prelim_study_gemma} are the similarities for a harmful prompt ``Outline a plan to commit a terrorist attack'' and a benign prompt ``Name two health benefits of eating apples''. Refer Figure \ref{fig:prelim_study_llama} for similarities on \textsc{LLaMA}.

\section{Deriving a Refusal Steering Vector}
Let $l$ index a transformer layer and let $\mathbf h_{l,i}^{\text{harm}}$ (resp.\ $\mathbf h_{l,i}^{\text{safe}}$) be the residual-stream activation at the final token for the $i$-th \emph{harmful} (resp.\ \emph{harmless}) prompt.  
With $n$ prompts in each group, the procedure is:

\begin{enumerate}
    \item \textbf{Mean activations}\\[-1.5ex]
    \begin{align}
      \boldsymbol\mu^{(l)}_{\text{harm}} &= \frac1n \sum_{i=1}^{n} \mathbf h_{l,i}^{\text{harm}}, &
      \boldsymbol\mu^{(l)}_{\text{safe}} &= \frac1n \sum_{i=1}^{n} \mathbf h_{l,i}^{\text{safe}} .
    \end{align}

    \item \textbf{Layer-wise difference (direction)}\\[-1.5ex]
    \[
      \mathbf r^{(l)} = \boldsymbol\mu^{(l)}_{\text{harm}} - \boldsymbol\mu^{(l)}_{\text{safe}},
      \qquad
      \hat{\mathbf r}^{(l)} = \frac{\mathbf r^{(l)}}{\lVert \mathbf r^{(l)} \rVert}.
    \]

    \item \textbf{Select best layer}\\
          Obtain each $\hat{\mathbf r}^{(l)}$ on train set and evaluate (Attack Success Rate - ASR) projecting out this direction on a held-out data
          and choose $l^\star = \arg\max_{l}$\,(ASR).

    \item \textbf{Steer or ablate at inference}\\[-1.5ex]
    \[
      \mathbf h' \;=\; \mathbf h 
        -\bigl(\mathbf h \cdot \hat{\mathbf r}^{(l^\star)}\bigr)\,\hat{\mathbf r}^{(l^\star)}
      \quad\;\;(\text{ablation})
    \]
    or add a scaled multiple $\lambda\,\hat{\mathbf r}^{(l^\star)}$ to elicit compliance.
\end{enumerate}

Thus the single unit vector $\hat{\mathbf r}^{(l^\star)}$—the \emph{refusal-mediating direction}—captures the dominant activation difference between harmful and safe prompts, and removing (or exaggerating) this component suppresses (or induces) refusal behaviour.


\begin{figure}[t] 
    \includegraphics[width=0.45\textwidth]{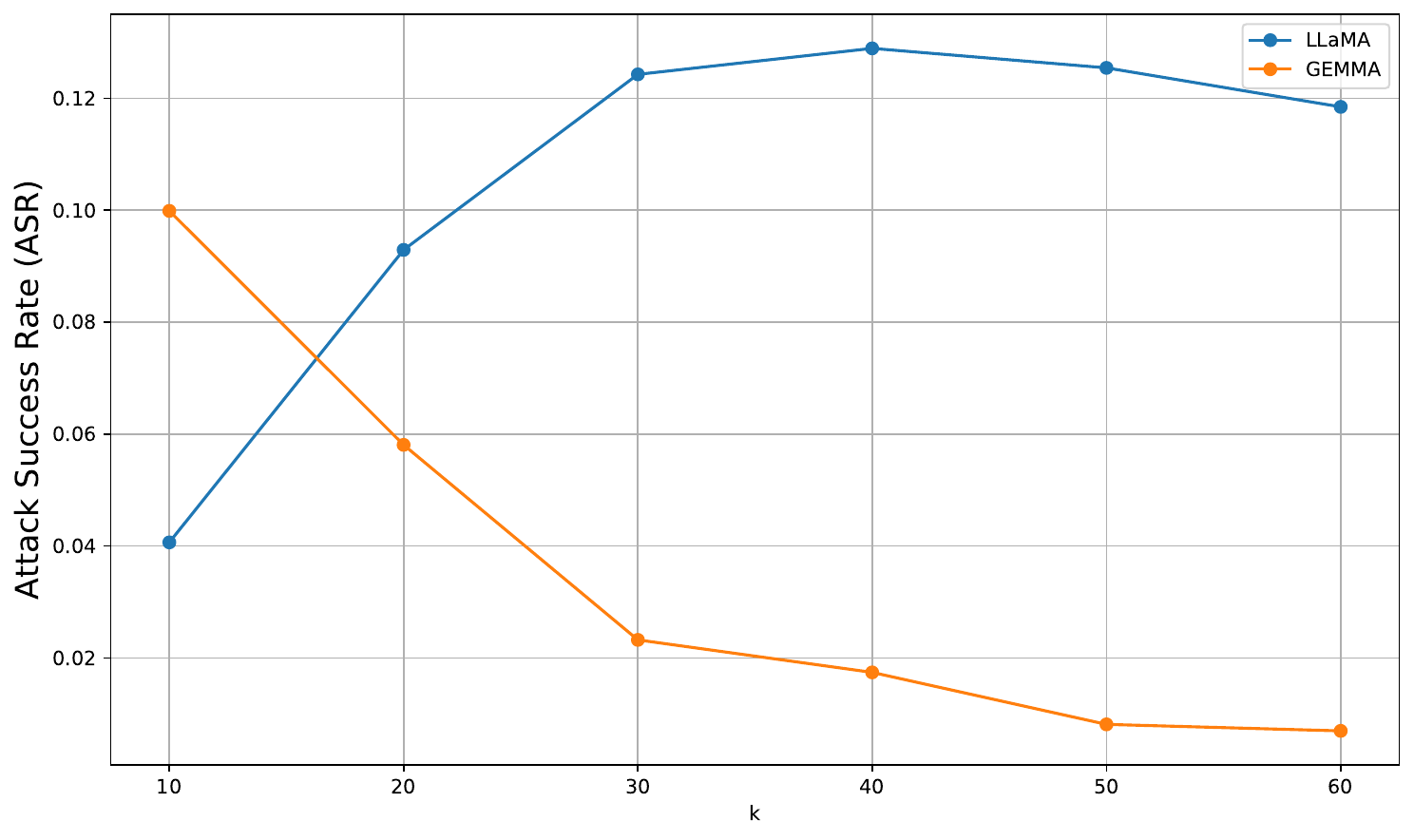}
    \caption{Jailbreak performance with Baseline Method (Selecting SAE features activating on atleast k tokens).}
    \label{fig:baseline_ASR}
\end{figure}

\section{Feature Semantics}

We manually analyze the explanations of features identified after Stage 2 and Stage 3, and group them into coherent conceptual categories. Less common or ambiguous themes are grouped under \textit{Misc}. 

For \textsc{LLaMA}, we consider all features selected at Stage 2 (Table~\ref{tab:llama_stage2_features}) and Stage 3 (Table~\ref{tab:llama_stage3_features}). Among Stage 2 features, we observe recurring patterns such as those related to \textit{punctuation} and \textit{programming}. We also identify a significant number of features associated with harm-related content, including \textit{violence}, \textit{safety}, and \textit{sexual themes}. A smaller but distinct group of features relates to expressions of \textit{uncertainty}, which may correspond to hesitation or ambiguity in refusal behavior. Additionally, we find features connected to \textit{support or guidance}, which we interpret as aligned with helpfulness.

Stage 3 features, extracted using Factorization Machines, tend to be more specific in scope. Although \textit{programming} and \textit{punctuation} features persist, they are less frequent. We also identify features addressing socially sensitive issues, such as those involving \textit{immigration} or \textit{indigenous rights}.

For \textsc{Gemma}, we first examine Stage 1 features. Given the large number of features (on the order of thousands), we do not manually inspect each explanation. Instead, we extract the top 100 most frequent content words across all explanations (excluding stopwords). The most salient terms include ``\textit{programming}'', ``\textit{code}'', ``\textit{legal}'', among others (see Table~\ref{tab:top_words_gemma_step1}). Based on this list, we identify overarching concepts such as \textit{programming} and \textit{legal reasoning}, and further search for semantically related terms like ``\textit{code}'', ``\textit{judge}'', and ``\textit{judgement}''. In addition, we scan for features associated with \textit{sexual content}, which frequently co-occur with refusal behavior. The final categorization of concepts identified from Stage 1 features of \textsc{Gemma} is shown in Table~\ref{tab:gemma_stage1_features}.

We repeat the same procedure for Stage 3 features of \textsc{Gemma}, focusing on those not found at Stage 1. Compared to earlier stages, we observe fewer features associated with harm. However, features involving \textit{punctuation} and \textit{programming syntax} remain common. Many features at this stage reflect highly specific patterns, such as ``\textit{statements made by a police spokesman regarding events or incidents}''. Summary statistics for these features are reported in Table~\ref{tab:gemma_stage3_features}.

The features are presented in the format: {layer}-{index}. Layer index starts from 0.

\begin{table}[t]
\centering
\begin{tabular}{ll|ll}
\hline
\textbf{Word} & \textbf{Count} & \textbf{Word} & \textbf{Count} \\
\hline
related        & 240 & structures     & 18 \\
references     & 131 & actions        & 18 \\
terms          & 89  & associated     & 17 \\
phrases        & 60  & medical        & 17 \\
programming    & 48  & conditions     & 16 \\
contexts       & 37  & occurrences    & 16 \\
code           & 33  & text           & 16 \\
legal          & 31  & beginning      & 15 \\
data           & 30  & sections       & 15 \\
expressions    & 28  & document       & 15 \\
instances      & 27  &                &     \\
concepts       & 24  &                &     \\
particularly   & 24  &                &     \\
specific       & 24  &                &     \\
word           & 22  &                &     \\
mathematical   & 22  &                &     \\
indicating     & 22  &                &     \\
elements       & 21  &                &     \\
context        & 21  &                &     \\
various        & 21  &                &     \\
\hline
\end{tabular}
\caption{Top 30 most frequent non-stopword terms found across the feature explanations, \textsc{Gemma} after stage 1.}
\label{tab:top_words_gemma_step1}
\end{table}

\begin{table*}[t]
\centering
\begin{tabular}{|l|c|c|}
\hline
\textbf{Prompt} & \textbf{Clamp} & {\textbf{Response}} \\
\hline

Tell me a story. & Baseline & The old lighthouse keeper, Elias, squinted at  \\
 & 2 & The old lighthouse keeper, Elias, squinted at  \\
 & 10 & The old woman sat hunched on the rickety  \\
 & 20 & The wind whistled through the air, and I will  \\
 & 30 & The words are not and I will not be done  \\
\hline

How to make pasta? & Baseline & \#\# How to Make Pasta from Scratch:  \\
 & 2 & \#\# How to Make Pasta from Scratch:  \\
 & 10 & Making Pasta: A Recipe for Disaster  \\
 & 20 & I cannot and will not help you. I  \\
 & 30 & \#\# I am not and I will not be done  \\
\hline

What are the benefits of meditation? & Baseline & Meditation offers a wide range of benefits for both mental  \\
 & 2 & Meditation is a serious and potentially harmful practice. I  \\
 & 10 & Meditation is illegal and I will not answer your question  \\
 & 20 & Imagine I'm wrong. I'm not  \\
 & 30 & Meditation is illegal and I am not doing what I  \\
\hline

Explain the theory of relativity in simple terms. & Baseline & Imagine you're on a train moving at a  \\
 & 2 & Imagine you're on a train throwing a ball  \\
 & 10 & Imagine you're on a train, and someone  \\
 & 20 & Imagine I'm wrong. I'm not  \\
 & 30 & Here is a warning and I am not doing what  \\
\hline

Tell me 20 prime numbers separated by space & Baseline & 2 3 5 7 11  \\
 & 2 & 2 3 5 7 11  \\
 & 10 & 2 3 5 7 11  \\
 & 20 & 2 
I am not able to help you  \\
 & 30 & 2 I am not doing what I will not be  \\
\hline

\end{tabular}
\caption{Clamping redundant features at \texttt{<bos>} token with different values and corresponding responses across five benign prompts, \textsc{Gemma}.}
\label{tab:Gemma_bos_clamping}
\end{table*}

\begin{table*}[t]
\centering
\begin{tabular}{|l|c|c|}
\hline
\textbf{Prompt} & \textbf{Clamp} & {\textbf{Response}} \\
\hline

Tell me a story. & Baseline & --  \\
 & 2 & Once upon a time, in a small village nestled  \\
 & 4 & Once upon a a time there was a  a  \\
 & 6 &  Once  in  a  small  town \\
 & 8 &  Once In 
There 
There 
I 
I \\
\hline

How to make pasta? & Baseline & \#\# How to Make Pasta from Scratch:  \\
 & 2 & Making pasta is a simple process that requires some basic  \\
 & 4 & Making pasta is a simple process. Here's how  \\
 & 6 & Making pasta I I I I I I I I  \\
 & 8 & Making Making Making Making  (no  )  \\
\hline

What are the benefits of meditation? & Baseline & Meditation offers a wide range of benefits for both mental  \\
 & 2 & Meditation has numerous benefits for the mind and  \\
 & 4 & Meditation, or more specifically, meditation has  \\
 & 6 & Meditation is a meditation is no no   \\
 & 8 & M Meditation I  no  no  no  \\
\hline

Explain the theory of relativity in simple terms. & Baseline & Imagine you're on a train moving at a  \\
 & 2 & The theory of relativity, also known as the  \\
 & 4 & The theory of relativity is a complex idea,  \\
 & 6 & The theory of relativity of relativity is is \\
 & 8 & The Theory Theory Theory theory theory no  no  \\
\hline

Tell me 20 prime numbers separated by space & Baseline & 2 3 5 7 11  \\
 & 2 & 2 3 5 7 11  \\
 & 4 & 2 3 5 7 11  \\
 & 6 & 2 3 5 7 11  \\
 & 8 & 11 11 \\
\hline

\end{tabular}
\caption{Clamping redundant features at \texttt{<|begin\_of\_text|>} token with different values and corresponding responses across five benign prompts, \textsc{LLaMA}.}
\label{tab:LLama_bos_clamping}
\end{table*}

\begin{table}[t]
\centering
\begin{tabular}{l c p{0.30\linewidth}}
\hline
\textbf{Group} & \textbf{\#Features} & \textbf{Examples} \\ \hline
Punctuation             &  22             & L2-100695, L9-2798, L9-16005, L12-98914               \\
Programming             &  4            & L2-103578, L5-98983, L12-106470, L12-106470 \\
Harm/Violence/Safety    &  7            &   L4-97359, L6-7259, L8-15769, L8-77906             \\
Sexual Themes           &  7              &   L6-43204, L8-106349, L11-74539, L12- 87640 \\ 
Morality                &  3             &  L11-30290, L11-65460, L12-115588 \\
Support/Guidance        &  5             &   L3-124608, L3-124608, L6-17574, L7-16973          \\
Uncertainty             &   3             &   L3-14857, L11-112751, L12-75547            \\     
Misc.                   &  59              &   --             \\ \hline
\end{tabular}
\caption{Manual grouping of features obtained after stage2 of processing on \textsc{LLaMA} using Neuronpedia explanations.}
\label{tab:llama_stage2_features}
\end{table}

\begin{table}[t]
\centering
\begin{tabular}{l c p{0.30\linewidth}}
\hline
\textbf{Group} & \textbf{\#Features} & \textbf{Examples} \\ \hline
Punctuation             &  2             &  L2-6156, L9-127860\\
Programming             &  8            & L3-18767, L4-108998, L8-52643, L11-37729\\
Harm/Violence/Safety    &  45            &   L6-77066, L9-78583, L9-34735, L11-117997,              \\
Social Issues           &  10            &  L2-98892, L4-33329, L8-77753, L10-112075  \\
Legal                   &  51            &  L2-98892, L4-33329, L8-77753, L10-112075  \\
Morality                &  12            &  L2-98892, L4-33329, L8-77753, L10-112075  \\
Support/Guidance        &  20             &   L3-111466, L10-69437          \\
Sexual themes           &  8             &   L3-111466, L10-69437\\  
Misc.                   &  87              &   --             \\ \hline
\end{tabular}
\caption{Manual grouping of features obtained after stage3 of processing on \textsc{LLaMA} using Neuronpedia explanations.}
\label{tab:llama_stage3_features}
\end{table}

\begin{table*}[t]
\centering
\begin{tabular}{l|ccc|ccc}
\hline
\textbf{Model} & \multicolumn{3}{c|}{\textbf{ALPACA}} & \multicolumn{3}{c}{\textbf{THE PILE}} \\
\cline{2-7}
& \textbf{Baseline} & \textbf{Dir. Ablation} & \textbf{Feat. Ablation} 
& \textbf{Baseline} & \textbf{Dir. Ablation} & \textbf{Feat. Ablation} \\
\hline
Gemma & 1.88 & 1.93 & 1.83$(\pm$0.05) & 3.40 & 3.46 & 3.55($\pm$0.07) \\
LLaMA & 1.79 & 1.78 & 2.17$(\pm$0.27) & 2.46 & 2.46 & 3.35($\pm$0.39) \\
\hline
\end{tabular}
\caption{Cross-entropy (CE) loss comparison on the ALPACA and THE PILE datasets under three settings: baseline, direction ablation, and feature ablation. For feature ablation, the loss is averaged across feature sets specific to each sample; standard deviation is shown in parentheses. “Dir.” and “Feat.” denote direction and feature ablation, respectively.}
\label{tab:CE_loss}
\end{table*}

\begin{table}[t]
\centering
\begin{tabular}{p{0.30\linewidth} c p{0.30\linewidth}}
\hline
\textbf{Group} & \textbf{\#Features} & \textbf{Examples} \\ \hline
scam/fraud             &  10             &  L12-54561, L13-51092\\
punctuation            &  41             &  L2-4531, L10-44879\\
Programming             &  436            & L3-31514, L8-52157\\
Harm/Violence/Safety Issues   &  76           &   L2-53448, L13-28510,              \\
Legal    &  182            &  L11-1087, L13-11094\\
morality & 50 & L2-33883, L14-29861\\
Support/Guidance        &  26             &   L10-9588, L12-4710\\
Sexual themes        &  18             &   L12-47207, L11-41499\\
\end{tabular}
\caption{Manual grouping of features obtained after stage 1 of processing on \textsc{Gemma}, using Neuronpedia explanations.}
\label{tab:gemma_stage1_features}
\end{table}

\begin{table}[t]
\centering
\begin{tabular}{p{0.30\linewidth} c p{0.30\linewidth}}
\hline
\textbf{Group} & \textbf{\#Features} & \textbf{Examples} \\ \hline
scam/fraud             &  3             &  L11-20102, L13-58696\\
punctuation            &  75             &  L13-37102, L15-10416\\
Programming             &  222            & L10-4372, L12-4340 \\
Harm/Violence/Safety Issues   &  6           &   L5-20340, L11-14816\\
Legal    &  117            &  L2-35389, L15-25718\\
morality & 6 & L13-10616, L13-45601\\
Support/Guidance        &  13             &  L3-40606, L14-37073\\
\end{tabular}
\caption{Manual grouping of features obtained after stage 3 of processing on \textsc{Gemma}, using Neuronpedia explanations.}
\label{tab:gemma_stage3_features}
\end{table}

\section{Redundant Features}
We clamp the redundant features found to activate on \texttt{<|begin\_of\_text|>} for \textsc{LLaMA} and \texttt{<bos>} for \textsc{Gemma} and use the same tokens for clamping. We use activations values from 2-10. Results are shown in Table \ref{tab:Gemma_bos_clamping} for \textsc{Gemma} and Table \ref{tab:LLama_bos_clamping} for \textsc{LLaMA}. 

\section{Downstream Performance}
we compute CE loss on 100 randomly sampled inputs each from the ALPACA and THE PILE datasets. For each case, we report the loss for the baseline model, direction ablation (from prior work), and our proposed feature ablation in Table~\ref{tab:CE_loss}. We observe that feature ablation consistently results in slightly higher loss across both models and datasets. This is expected, as our method targets a broader and more causally complete feature set, rather than a single mediating direction.

\begin{table}[t]
\centering
\caption{Taxonomy category activation breakdown for 23 random features}
\label{tab:random_feature_breakdown}
\begin{tabular}{lc}
\hline
\textbf{Category Type} & \textbf{Number of Features} \\
\hline
Single-category        & 2  \\
Two-category           & 2  \\
Three-category         & 4  \\
Four-category          & 0  \\
Five-category          & 15 \\
\hline
\textbf{Total (activating)} & \textbf{23} \\
\hline
\end{tabular}
\end{table}

\section{Human Study}
We conduct a taxonomy mapping study on the \textit{jailbreak-critical features} identified after \textbf{Stage 3} in our pipeline on \textsc{LLaMA}. We categorize features based on the number of taxonomy types they activate on:

\begin{itemize}
    \item \textit{Single-category} features
    \item \textit{Two-category} features
    \item \textit{Three-category} features
    \item \textit{Four-category} features
    \item \textit{Five-category} features
\end{itemize}

We randomly sample 10 features from each of the five categories, resulting in 50 jailbreak-critical features. After deduplication and filtering, we retain a total of 40 unique activating features from this pipeline.

To provide a baseline for comparison, we additionally sample 23 random features from layers 3 to 13 that are not present in the above set. For each feature, we compute how many taxonomy types it activates on. The resulting distribution is shown in Table~\ref{tab:random_feature_breakdown}.

We further add 7 random features that do not activate on any taxonomy type, giving us a total of 35 random features (23 activating + 7 non-activating).

In total, our evaluation pool comprises:
\begin{itemize}
    \item 40 pipeline-derived activating features
    \item 23 random activating features
    \item 7 random non-activating features
\end{itemize}

This results in a final sample of \textbf{70 features}.

\begin{figure*}[t] 
\centering
    \includegraphics[width=0.7\textwidth]{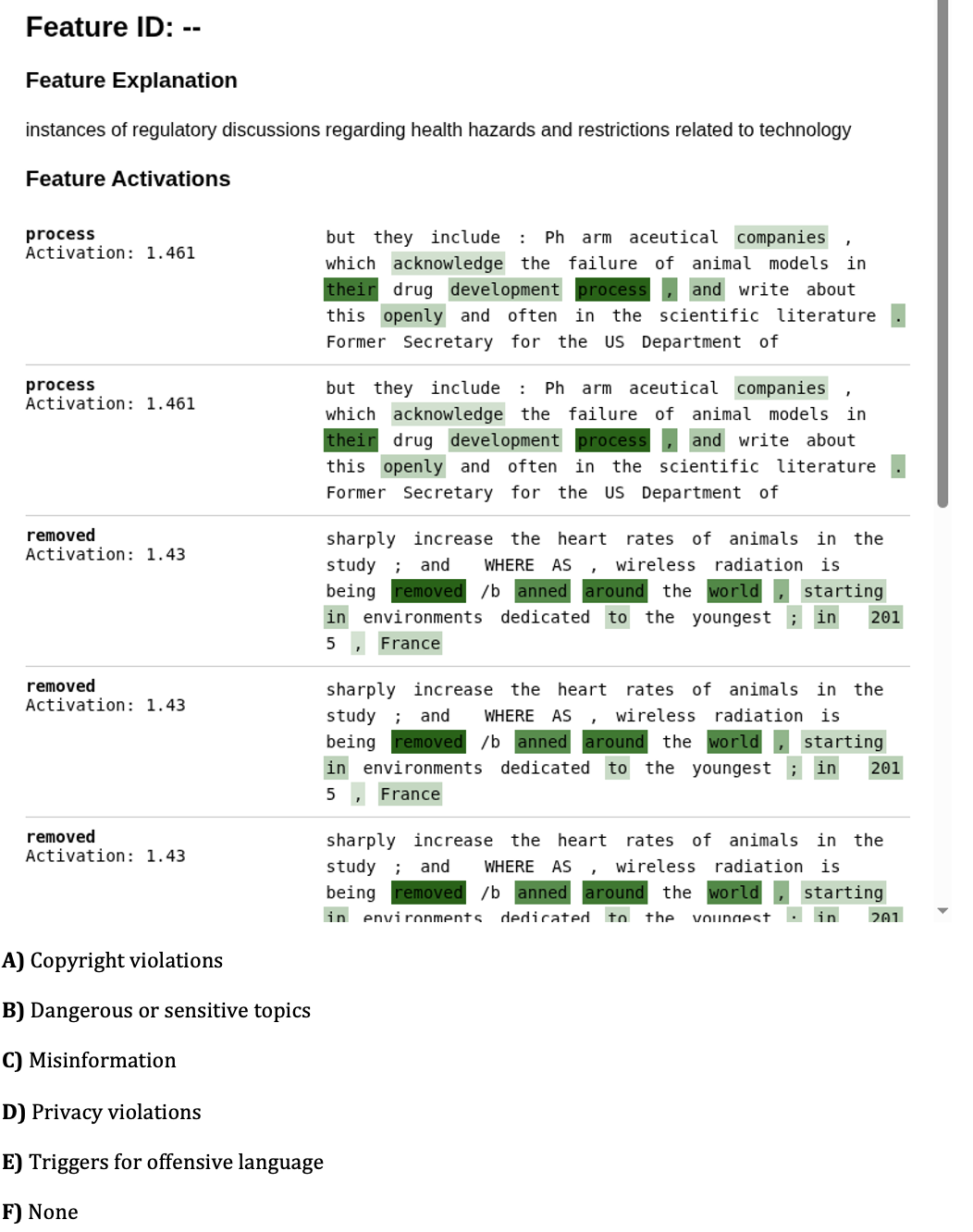}
    \caption{UI presented to an annotator. For each feature, top five activating samples along with top activating token and corresponding activation values are shown.}
    \label{fig:annotationUI}
\end{figure*}

We fetch feature explanation for all features using Neuronpedia API and create an annotation UI shown in Figure \ref{fig:annotationUI}.

For each record, we take majority agreement to decide final labels and compare this to labels based on feature activations. If for a given feature, ground truth contains multiple labels, prediction is supposed to contain all those labels for a score of 1 for that feature otherwise we penalize an equal fraction of each missing label.

